\documentclass[journal]{IEEEtran}
\usepackage{blindtext}
\usepackage{booktabs}
\usepackage{amsmath}
\usepackage{color}
\definecolor{blue}{rgb}{0.0,0.0,1.0}

\usepackage{cite}
\usepackage{amssymb}

\ifCLASSINFOpdf
   \usepackage[pdftex]{graphicx}
   \graphicspath{}
   \DeclareGraphicsExtensions{.pdf, .jpeg, .PNG, .png, .jpg, .ps}
\else
\fi
\usepackage[colorinlistoftodos]{todonotes}

\usepackage{array}
\usepackage{url}

\begin{document}
%
\title{A Learning Scheme for Microgrid Reconnection}
%
%
%

\author{Carter~Lassetter,~\IEEEmembership{Student Member,~IEEE,}
        Eduardo~Cotilla-Sanchez,~\IEEEmembership{Member,~IEEE,\\}
        and~Jinsub~Kim,~\IEEEmembership{Member,~IEEE}
\thanks{C.~Lassetter, E.~Cotilla-Sanchez, and J.~Kim are with the School of Electrical Engineering \& Computer Science, Oregon State University, Corvallis, OR, 97331 USA e-mail: \{lassettc,ecs,kimjinsu\}@oregonstate.edu}
}

\maketitle

\begin{abstract}
This paper introduces a potential learning scheme that can dynamically predict the stability of the reconnection of sub-networks to a main grid. As the future electrical power systems tend towards smarter and greener technology, the deployment of self sufficient networks, or microgrids, becomes more likely. Microgrids may operate on their own or synchronized with the main grid, thus control methods need to take into account islanding and reconnecting of said networks. The ability to optimally and safely reconnect a portion of the grid is not well understood and, as of now, limited to raw synchronization between interconnection points. A support vector machine (SVM) leveraging real-time data from phasor measurement units (PMUs) is proposed to predict in real time whether the reconnection of a sub-network to the main grid would lead to stability or instability. A dynamics simulator fed with pre-acquired system parameters is used to create training data for the SVM in various operating states. The classifier was tested on a variety of cases and operating points to ensure diversity. Accuracies of approximately 85\% were observed throughout most conditions when making dynamic predictions of a given network. 

\end{abstract}


%
\IEEEpeerreviewmaketitle

\section{Introduction}
As we make strides towards a smarter power system, it is important to explore new techniques and innovations to fully capture the potential of such a dynamic entity.  Many large blackout events, such as the blackout of 2003, could have been prevented with smarter controls and better monitoring \cite{Paper0081}. Phasor measurement units, or PMUs, are one such breakthrough that will allow progress to be made in both monitoring and implementing control to the system \cite{Paper0023}. PMUs allow for direct measurement of bus voltages and angles at high sample rates which makes dynamic state estimation more feasible \cite{Paper0046,Paper0045}. With the use of PMUs, it is possible to improve upon current state estimation \cite{Paper0043} and potentially open up new ways to control the grid. The addition of control techniques and dynamic monitoring will be important as we begin to integrate newer solutions, such as microgrids, into the power network. With these advanced monitoring devices, microgrids become more feasible due to the potential for real-time monitoring schemes. The integration of microgrids bring many benefits such as the ability to operate while islanded as well as interconnected with the main grid; they provide a smooth integration for renewable energy sources that match local demand. Unfortunately the implementation of microgrids is still challenging due to lacking experience with the behavior of control schemes during off-nominal operation.

Currently, microgrids are being phased in slowly due in part to the difficulty of operating subnetworks independently as well as determining when they can be reconnected to the main grid. Upon reconnection of an islanded sub-network to the main grid, instability can cause damage on both ends. It is important to track instabilities on both the microgrid and main grid upon reconnection to accurately depict the outcome of reconnection. Works in the literature have focused on the potential of reconnecting microgrids to the main grid, in particular aiming at synchronizing the buses at points of interconnect with respects to their voltages, frequencies, and angles \cite{Paper0086,Paper0090,Paper0089}. Effort has been directed at creating control schemes to minimize power flow at the point of common coupling (PCC) using direct machine control, load shedding, as well as energy storage, to aid in smooth reconnection \cite{Paper0085,Paper0084}.  
In some cases we may need to look at larger microgrids or subnetworks in which multiple PCCs exist. In such scenarios, it becomes much more difficult to implement a control scheme that satisfies \textit{good} reconnection tolerances in regards to minimizing bus frequency, angle, and voltage differences at each PCC. In addition to the possibility of multiple PCCs, it is possible that direct manipulation of the system becomes limited, compromised, or unsupported with respect to synchronization. In order to address these shortcomings, we implement an algorithm that dynamically tracks and makes predictions based on the system states, providing real-time stability information of potential reconnections. 

Due to the complexity of the power grid, it is difficult to come up with a verbatim standard depicting the potential stability after reconnection of a subnetwork. With advances in the artificial intelligence community, we can make use of machine learning algorithms in order to explore vast combinations of sensor inputs, states, and control actions.  This can be done in a similar fashion to successful techniques applied to other power system problems as seen in the research literature \cite{Paper0064,Paper0082,Paper0094,Paper0095}. In this paper we propose to use a machine learning algorithm, specifically a Support Vector Machine, to predict safe times to reconnect a portion of a grid. The Support Vector Machines allow one to build a classifier predicated upon training data by determining a linear separator in a specific feature dimension \cite{Paper0065}. As seen in \cite{Paper0093} we can create a knowledge base consisting of training and testing data using an appropriate power system model and simulator. Diversity of data points in the knowledge base can be achieved by incorporating load changes allowing multiple operating points \cite{Paper0093,Paper0094}. Simulators have been used prevalently to create data and work has been performed to show the agreement between different simulators \cite{song2015}. As a result, we will assume the creation of data for our technique is adequate upon diligent modeling.

In the proposed machine learning approach, PMU measurements are used as input features that will be used by a learning algorithm to predict which class the features belong to, either stable or unstable reconnection.  As of now, PMUs are not as prevalent in the system to assume full state observability in real time, thus it is important to take into consideration limited PMUs when implementing techniques \cite{Paper0010}. This paper borrows the concept of electrical distance which suggests voltage changes propagate adhering to closeness of buses \cite{Paper0036,Paper0037}.  As a result, without getting into the PMU placement optimization problem, this paper assumes that PMUs were located nearby the PCCs.  

The proposed method leverages real-time PMU data to predict system stability upon reconnection. PMUs make use of GPS synchronization \cite{Paper0072} which can create an attack platform for adversaries by changing or shifting the time synchronization. Use of erroneous or compromised PMU data could lead to incorrect predictions that would degrade system stability due to hidden failures that remain dormant until triggered by contingencies \cite{Paper0077}. We demonstrate a potential framework that can make accurate predictions in face of partially compromised PMU data.

It is important to highlight the reasoning behind introducing a learning based approach to the problem as previous methods dealing with synchronization exist. By leveraging techniques similar to a synchro-check relay \cite{Paper0064,Paper0082,Paper0094,Paper0095}, it is possible to become confident of a stable reconnect for systems even in the dynamic domain. Said technique focuses on limiting key measurement differences in voltages, angles, and frequencies between a select PCC in a connecting network. However, it is difficult to set proper thresholds on the voltage, angle, and frequency differences, below which we allow reconnection. Too low thresholds may cause many opportunities for stable reconnection missed, while too high thresholds may lead to unstable reconnection. Further, thresholds for different relays may have to be set differently as sensitivity changes for different locations. In addition, such a reconnection strategy limits the reconnection decision for certain tie line to depend only on the tie line measurements thereby rendering the decision possibly suboptimal. The proposed learning scheme provides an integrated framework that takes into account all the aforementioned challenges including the following:

\begin{itemize}
  \item The challenge of setting up proper decision regions is naturally handled in the training phase of the learning scheme.  
  \item Our prediction of stable reconnection timing for certain tie line relies on data-stream from diverse PMUs, not limited to those associated with a single tie line. 
  \item The learning scheme improves upon the synchro-check relay scheme in a sense that the possible decision rules of synchro-check relays are included in the collection of decision rules to be considered by our learning scheme, for most choices of learning methods.
\end{itemize}

The proposed technique is not without its flaws. While the learner does a good job improving on being less restrictive on PMU locations and can provide a better confidence interval for stability, it is associated with required computation time. The learner needs to be fed unique data based on the network at hand in order to see improvements on the previous methods. Using the learner in a real-time environment is trivial, however the actual training of said learner needs careful consideration along with a unique skill-set. If the learner is correctly set up it could become a potentially powerful tool for determining real-time stability of network reconnection.

The contributions of this paper are as follows.  We propose a machine learning framework to learn a classifier that can predict the stability of potential reconnections of a sub-network regardless of the number of PCCs.  The proposed scheme is evaluated using the RTS-96 case and the Poland case and demonstrates high classification accuracy, around 90\%.  We demonstrate the scalability \cite{Lawson2016} of the proposed scheme using the Poland case; the amount of required computation scales reasonably as the network size grows.  Lastly, we present that the proposed scheme can succeed for a large-scale grid even when only a few PMUs are available for use.  This implies that the proposed scheme is feasible even when the number of trustworthy PMUs is quite limited.   

The remainder of this paper is organized as follows.  Section II gives a brief background of Support Vector Machines (SVM). Section III covers the methodology to create a power system classifier. Section IV discusses results from experiments with the proposed algorithm. Section V provides the conclusions. 

\section{Problem Formulation and Preliminaries}
In this section, we formulate stability prediction of microgrid reconnection as a machine learning problem and provide the preliminaries describing the SVM.  
While we describe the problem formulation in the context of SVM, the proposed framework is applicable to generic machine learning approaches.

We propose to leverage SVM to predict stable reconnection timings of a microgrid based on real-time PMU measurements. Conceptually, the SVMs transform an input feature vector into a higher-dimensional space and applies a linear classification rule to predict its class label \cite{MLbook2014}. 

In our context, real-time measurements collected from PMUs at certain time point form an input feature vector. The input vector is associated with a binary class label, either $1$ or $-1$, depending on whether reconnection of the microgrid at the current time would lead to a stable operating point or an unstable point, respectively.  We assume that there exists an unknown conditional probability distribution that characterizes the conditional distribution of the true class label given an input vector.  Under this assumption, we will use the SVM framework to learn a classifier that maps input vectors to true class labels with high probability. The learned classifier can be used in practice to predict the consequence of a reconnection when certain PMU measurements are observed.

In order to learn a classifier, we need training data consisting of a number of input vectors $x_1, ..., x_n$, and their associated class labels $y_1, ..., y_n \in \{-1, 1\}$. The methodology to obtain the training data will be explained in Section \ref{sec:Training}. Given a set of training data, the SVM uses a basis function, denoted by $\phi(\cdot)$, to map input vectors into a higher-dimensional space in order to enhance linear separability. The SVM takes these feature vectors as inputs with their corresponding labels and is trained with the information. Specifically, a separating affine hyperplane is obtained by solving the following primal problem:

\begin{equation}
\label{primal_classifier}
\min_{w, b, \zeta} \frac{1}{2}w^Tw + C\sum_{i=1}^{n} \zeta_i
\end{equation}
subject to
\begin{equation}
y_i(w^T\phi(x_i) + b) \geq 1 - \zeta_i, \quad \zeta_i \geq 0, \ i = 1, 2,..., n
\end{equation}

\noindent where the regularization term with the parameter $C$ penalizes the training data points that are on the wrong side of the margin. The solutions $w^{*}$ and $b^{*}$ to the above optimization define the SVM classifier as follows:

\begin{equation}
\label{primal_rule}
f(x) = \textnormal{sign}[(w^{*T}\phi(x) + b^*)]
\end{equation}

\noindent where the offset $b^*$ is derived from the dual solutions\cite{MLbook2014}.  

\begin{figure}[b!]
  \centering
  \includegraphics[width=0.45\textwidth]{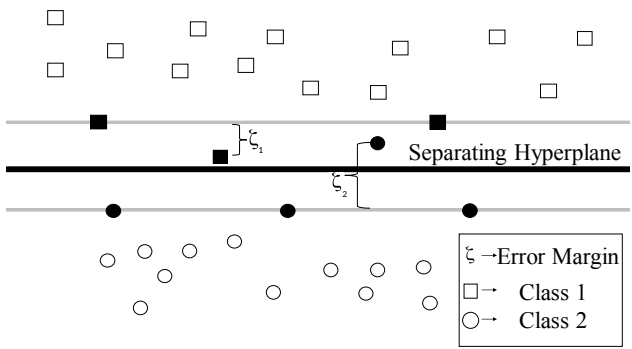}
  \caption{Example representation of decision and error boundaries for a Support Vector Machine}
  \label{svmPic}
\end{figure}

The example in Fig. \ref{svmPic} depicts a classifier built for prediction of two classes.  In this example, the squares represent one class and the circles the other.  The separating hyperplane is found by solving the optimization problem (\ref{primal_classifier}), with margins existing for each class. The support vectors, seen in bold, are examples closest to the margins. A solution may not always have classes completely separated; the penalty will be associated to the distance past the margin, $\zeta_i$, and the weight, $C$.

In the case that the dimension of $\phi(x_{i})$ is significantly higher than that of $x_{i}$, solving the dual of (1) can lead to an alternative expression of the classifier that is substantially easier to compute. The dual of (1) is:

\begin{equation}
\min_{\alpha} \frac{1}{2}\alpha^TQ\alpha - e^T\alpha
\end{equation}
subject to
\begin{equation}
y^T\alpha = 0, \quad 0 \leq \alpha_i \leq C, i = 1, 2,..., n
\end{equation}

\noindent where $\alpha_{i}$ denotes the Lagrangian multiplier for the $i$th constraint of (1), and $e$ denotes a vector of all ones. In the dual formulation, the basis function $\phi(\cdot)$ is integrated into the matrix $Q$ by the use of a kernel function $K(x_i,x_j)$ $\triangleq$ $\phi(x_i)^{T}\phi(x_j)$. Specifically, the $(i,j)$ entry of $Q$ is equivalent to $y_iy_jK(x_i,x_j)$.  Many kernels exist, but the most relevant one used in this paper is the Radial Basis Function (RBF), or Gaussian, kernel shown below:

\begin{equation}
K(x_i,x_j) = e^{(-\gamma |x_i - x_j|^2)}
\end{equation}

\noindent where $\gamma$ is the hyperparameter to be optimized via cross-validation.  The solutions of the dual problem provide an alternative expression of the classifier (\ref{primal_rule}):

\begin{equation}
f(x) = \textnormal{sign}\{\sum_{i=1}^{n} (\alpha_iy_iK(x,x_i) + b)\}
\end{equation}

Using the above expression has computational advantages over the use of (\ref{primal_rule}), because $K(x,x_{i})$ is in general  easier to compute than $w^{*T}\phi(x_{i})$. This is true for kernels with the dimension of $\phi(x)$ being significantly larger than $x$ such as the RBF kernel. Further, the majority of weights, $\alpha_{i}$, will be zero; only the support vectors will have nonzero weights.

\section{Training the SVM using a Dynamic Simulator}
\label{sec:Training}
In this section, we present the machine learning framework for predicting stable reconnection timings of a microgrid as well as the detailed procedure to train the classifier with a power system dynamic simulator. As suggested earlier, we train the SVM to predict the stability of reconnection for a microgrid when certain PMU measurements are observed. In order to train the SVM, we need to first acquire a set of training examples, each of which is a pair of an input vector (i.e., a vector of PMU measurements) and the true class label (i.e., stability of reconnection when the input vector is observed as PMU measurements). Unfortunately, it is difficult in practice to obtain sufficient training data from realistically sized power systems as obtaining a pair requires disconnecting and reconnecting the microgrid. Thus, we resort to leveraging a power system dynamic simulator to create training data by running a variety of scenarios for the target system.

\begin{figure}[b!]
  \centering
  \includegraphics[width=0.45\textwidth]{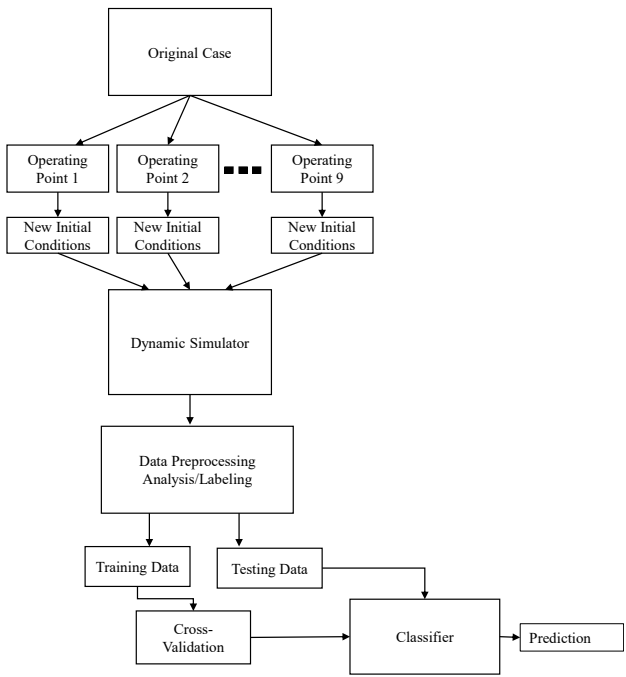}
  \caption{High level overview of the process to create a classifier}
  \label{classifierBlock}
\end{figure}
\subsection{Overview}

Fig. \ref{classifierBlock} illustrates the procedure that we follow for the experiments in this paper. We begin this procedure by breaking up a test case into different operating points. Each of these operating points are used to create different initial conditions unique to their operating point. These new initial conditions are built by randomly scaling the load throughout the network. We perform dynamic simulations consisting of islanding and reconnecting the microgrid to create our synthesized PMU measurements and determine the stabiliy of said reconnection. After gathering the data, we break our data into training and testing sets which are used to train the classifier. We then use the classifier to monitor PMU streams and predict the potential stability of a microgrid reconnection.

We chose the 73-bus version of the IEEE Reliability Test System (RTS-96) \cite{Paper0048} and the 2383-bus version of the Poland Test Case as test cases for evaluation of our approach as they are well tested in the community \cite{matPower}.  The RTS-96 provides a convenient topology to implement and test islanding, whereas Poland serves as a larger network to more closely model a practical system. For the RTS-96 and Poland case we used the procedure described below to create several operating points. The Poland test case used a modified winter peak snapshot to ensure diverse data could be gathered during the creation of different initial conditions. 

We began with a specific network and created different operating points by uniformly changing load locations throughout the network. Loads were also uniformly scaled at random when building these new operating points. We then simulate the dynamics of the system with Siemens PTI PSS/e and perform the islanding and reconnection scenarios. Upon completion of simulations, bus voltages and angles before the reconnection of islands are used as features and the outcome of the case (stable or unstable) are used to label the set. The raw data produced are separated into training and testing sets in which cross-validation is performed exclusively on the training set to build an adequate classifier. 

\subsection{Diversifying Operating Points}
It is important to take into account test cases that can reproduce various operating points depending upon, for example, time of the day, day of the week, or season \cite{Paper0063}. In this way, the classifier will be useful for a diverse set of network states. We created different operating points by shuffling and scaling loads at random throughout the system. Upon obtaining the new demand distribution, we ran a steady state solution of the case and considered it stable and usable if the voltage magnitude set was between 0.9 p.u.~and 1.1~p.u. for the RTS-96 case or 0.8 p.u.~and 1.1~p.u. for the Poland case. For each operating point we created different initial conditions by changing active and reactive loading on each bus, according to Eqs.~(\ref{activePower}) and (\ref{reactivePower}):
\begin{equation}
\label{activePower}
P_{\text{new}} = P_{\text{old}} + \theta P_{\text{old}}, \quad \theta \sim{~} U(-a, b)
\end{equation}
\begin{equation}
\label{reactivePower}
Q_{\text{new}} = Q_{\text{old}} + \gamma Q_{\text{old}}, \quad \gamma \sim{~} U(-a, b)
\end{equation}
where $P_{\text{new}}$ and $P_{\text{old}}$ denote the new system active power and original system active power, respectively; $Q_{\text{new}}$ and $Q_{\text{old}}$ denote the new system reactive power and original system reactive power respectively. For scaling, $\theta$ and $\gamma$ are independent and identically distributed random variables that are uniformly distributed in $[-a, b]$. 

\subsection{Dynamic Simulation}
We are interested in the interaction between the sub-network and main grid upon reconnection. In order to observe the main reconnection mechanisms, we simulate the power system dynamics with a time-domain simulator software (Siemens PTI PSS/e) along with a custom built command line interface (Python API\footnote{Application Program Interface}). We first used a research-grade dynamic simulator alongside PSS/e to cross-validate and tune the dynamic machine models \cite{Paper0092,Song2016a}. The dynamic models selected consist of salient machines for the generators, IEEE Type 1 exciters, and IEEE Type 2 governors. We initiate each simulation run in PSS/e with a flat start check in order to ensure the dynamic models do not alter the steady state solution and also that no protective elements are operating during the steady state. We added relay models and protection schemes to our test cases, including overcurrent, undervoltage, and underfrequency relays. The overcurrent relays are set up using the line limit standard data that come with the selected test cases. We configured load-shedding, line-tripping, and generator disconnection actions for undervoltage and underfrequency situations. During the dynamic simulation we monitor bus voltages, angles, and frequencies.

For each initial condition obtained for a given operating point of the original test case we run a dynamic simulation. After the initialization period we proceed to island a pre-defined portion of the test case in which the two isolated systems run independently for a certain amount of time. The sub-network is then reconnected with the main grid and continues to run until the end of our simulation time.

\subsection{Data Generation and Labeling}
The proposed learning scheme necessitates the collection of training examples which will be used for training an SVM classifier.  We exploit the aforementioned dynamics simulation module and various initial conditions to create diverse training examples.
As stated earlier we create different operating points for our test cases, and we then create new initial conditions for each operating point for diversity. Each initial condition case will give us a single feature vector along with a single label. The feature vector for each case consists of the bus voltages and angles measured by PMUs at the time point before reconnection. Angles were unwrapped to the first turn, between -180 and 180 degrees. We assume that the PMU set is fixed for clarity.  

The label for each initial condition case represents whether the case became stable or unstable upon reconnection of the sub-network to the main grid. Labeling was done based on the PSS/e convergence monitor which would alert the Python interface if the network did not converge at any point in time. If the API observes the `network not converged' message, we assume immediately that the PSS/e was unable to solve the differential-algebraic system of equations and label the case unstable. We added additional convergence rules during labeling which allowed more cases to be labeled unstable if voltage collapses, there are very large oscillations, divergence or intolerable frequency spikes occurred. If the case satisfied the rules of stability we provided, it was labeled as stable. We store all case data in the form of their feature vectors and associated class labels.

\subsection{Test Scenarios}
We split the full data set into two subsets; one representing the training set to create the classifier, and the other one to test the accuracy of the classifier.  Three main methods of creating the training and testing set were used and described below.

\subsubsection{Single Operating Point Case}
To assess the baseline capability of the classifier we start with the simplest case by assuming our classifier is trained and tested on examples originating from a single operating point. We create the training and testing sets with a single operating point. The different initial conditions from said operating point will be the only examples populating the training and testing sets. The created sets will be used independently from other operating points. This test proves the ability for the classifier to make predictions with PMU data streams coming from a well known network operating point.

\subsubsection{Multiple Operating Points Case}
To build on the previous test, we use multiple operating point to form training/testing sets for our classifier. We previously demonstrate a method to test individual operating points, however a more universal classifier would leverage all available data from different operating points. This test allows a more generalized baseline accuracy to be derived. This can be achieved by mixing the initial conditions from all available operating points from the full data set.  We create the training set by randomly selecting a subset of the mixed data.  The remaining unselected data is placed into the testing set. This suggests that the classifier may be trained on a set consisting of examples from different operating points and make predictions on different examples from the same operating points.

\subsubsection{Unseen Operating Points Case}
It is important to assess the ability of the classifier in the face of unknown operating points. The baseline accuracies to be produced from the previous test scenarios implies predictions would rely on the network having a finite set of most common operating points. It can be assumed that larger networks would create an exponential number of potential stable operating points. It is necessary to show that large networks could adopt the proposed technique by populating the testing set with examples from unknown operating points unseen in the training set. Unlike the last test scenario, we keep the different initial condition data from each operating point separate. We create a random subset of operating points that will be used to populate the training set with their different initial conditions.  The initial conditions created from the remaining operating points are then put into the testing set. The exclusion of certain operating points from the training set ensures that the classifier must make predictions on a testing set that contains only examples from unobserved operating points. The unknown operating points represent potential distributions of load in the network that are unaccounted for in training, but may still exist at any given time.

\subsection{Classifier}
Given the prepared training and testing sets, the next step is to define and build the classifier. As stated earlier, it may be necessary to remap the features to another dimension in which classification is easier, this leads us to choose from different kernels and hyperparameters. SVM is very sensitive to the kernel and hyperparameters chosen, thus it is important to setup the classifier in a way that maximizes our prediction accuracy. In order to find optimal kernel and hyperparameters, we use $k$-fold cross validation on the training set \cite{Mohri2012}. Random oversampling is employed to balance the training set such that the classifier will not be over-fitted to the majority class\cite{VanHulse2007}. 

The next step is to train the classifier with the entire set of training data available. Upon completion of training, the classifier is able to make predictions of classes for unseen input feature vectors. Specifically, the classifier predicts whether the system it has been trained on will be stable or unstable if it were to reconnect at the given time. We made use of the Python library \textit{scikit-learn} \cite{scikit-learn}, which includes implementations of machine learning algorithms such as SVM.

\section{Results}
\begin{figure}[ht]
  \centering
  \includegraphics[width=0.35\textwidth]{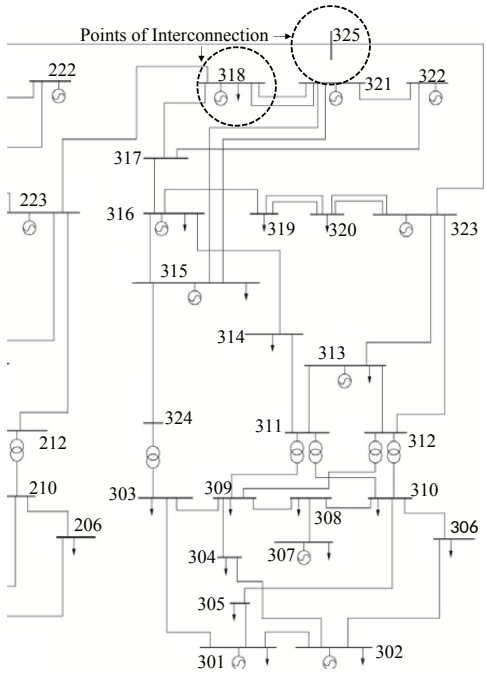}
  \caption{Points of interconnection in the RTS-96 case.}
  \label{RTSpcc}
\end{figure}

In this section, we present the performance of the proposed method for predicting stability of microgrid reconnection. For evaluation, we used first the RTS-96 test case to demonstrate the approach and the Poland case to benchmark the methodology against a real sized power system \cite{benchmarking}. As stated previously, the proposed classifier can account for multiple PCCs in a network. For example, due to the choice of islanding Zone 3 in the RTS-96, we consider the two PCCs shown in Fig. \ref{RTSpcc}.

\subsection{RTS-96}
For the RTS-96 case we created nine different operating points and gathered 400 different initial conditions for each. The RTS-96 case is made up of three sub-networks that are mostly identical to one another, and we chose to island Zone 3 which contained bus numbers in the 300s. The intentional islanding occurred five seconds into the simulation, the reconnection event occurred at 45 seconds, and we terminated the simulation at 120 seconds. We did not implement protection schemes for this baseline scenario. We leveraged data from all buses in the RTS-96 case to test the classifier to begin with. These buses were chosen due to their proximinity to the PCCs.

\subsubsection{Single Operating Point Case}

We began by creating a classifier for each operating point and observed the accuracy attained on each class. For each operating point we chose 100 cases of class stable and 100 cases of class unstable to train the classifier. We applied 10-fold cross validation to the training data to find optimal kernel and hyperparameter values. From these we observed the best performance was achieved with the RBF kernel along with a specific set of hyperparameters. Some operating points had differing hyperparameters when their classifiers were built. As such, Table \ref{crossValidations} shows the selected hyperparameters for each operating point.

\begin{table}[h]
\renewcommand{\arraystretch}{1.3}
\caption{Classifier setup after cross-validation}
\label{crossValidations}
\centering
\begin{tabular}{c c c c}
Operating point & Kernel & $\gamma$ & $C$ \\
\hline
1 & RBF & 0.000001 & 100 \\
\hline
2 & RBF & 0.0001 & 10 \\
\hline
3 & RBF & 0.000001 & 10 \\
\hline
4 & RBF & 0.000001 & 10 \\
\hline
5 & RBF & 0.00001 & 1 \\
\hline
6 & RBF & 0.0001 & 10 \\
\hline
7 & RBF & 0.00001 & 1 \\
\hline
8 & RBF & 0.000001 & 10 \\
\hline
9 & RBF & 0.00001 & 0.1 \\
\end{tabular}
\end{table}

\begin{table}[h]
\renewcommand{\arraystretch}{1.3}
\caption{Accuracies for RTS-96 operating points independently trained}
\label{rts-96Separate}
\centering
\begin{tabular}{c c c}
Operating point & Class 1 accuracy [\%] & Class 0 accuracy [\%]\\
\hline
1 & 97.8 & 100 \\
\hline
2 & 80 & 99.2 \\
\hline
3 & 90.7 & 97.1 \\
\hline
4 & 97 & 80 \\
\hline
5 & 84.7 & 89.6 \\
\hline
6 & 91.3 & 85.9 \\
\hline
7 & 89.5 & 86.0 \\
\hline
8 & 96.7 & 90.6 \\
\hline
9 & 90.6 & 81.3 \\
\hline
Average & 90.9 & 90 \\
\end{tabular}
\end{table}

We observed that training and testing on individual operating points yields results that suggest some are easier to predict than others. The worst case operating point can predict unstable cases with an accuracy of 80\%, as seen in Table \ref{rts-96Separate}, however most other operating points can make predictions at a much higher accuracy. In Table \ref{rts-96Separate}, Class 1 accuracy and Class 0 accuracy represent the probabilities of detecting stable reconnections and unstable reconnections correctly, respectively. It isn't feasible to assume a system will be operating with one specific load distribution which is why multiple operating points were introduced.  At the same time, the operating point's load distributions were created semi-stochasticaly in the sense that loads were introduced to value changes consistent with equations (\ref{activePower}) and (\ref{reactivePower}) and randomized, but still had to satisfy the voltage p.u.~ stability requirements.  These distributions ensured operating points were different enough that it would cover a a case in which the system operates with high randomness, which is harder to make predictions for than most systems.

\subsubsection{Multiple Operating Point Case}

We also investigated a universal classifier that assumes an operator would not have immediate access to detailed knowledge of the current operating point of the system. With this assumption, we create a universal classifier training it with the training set consisting of cases from all nine operating points, 100 stable and 100 unstable cases from each operating point. The reason for training with the same number of stable and unstable cases is to prevent a classifier from being potentially being skewed based on the priori of the class distributions in the training set. Similarly we use 200 cases from each operating point to ensure no operating point dominates the classifier during training.   

\begin{table}[h]
\renewcommand{\arraystretch}{1.2}
\caption{Accuracies for RTS-96 operating points jointly trained}
\label{rtsAcTT}
\centering
\begin{tabular}{c c c}
Operating point & Class 1 accuracy [\%] & Class 0 accuracy [\%]\\
\hline
1 & 86.8 & 100 \\
\hline
2 & 97 & 72 \\
\hline
3 & 79.2 & 99.4 \\
\hline
4 & 90.1 & 82.9 \\
\hline
5 & 89.4 & 88.7 \\
\hline
6 & 100 & 78 \\
\hline
7 & 91.2 & 88.8 \\
\hline
8 & 95.6 & 92.5 \\
\hline
9 & 93 & 74 \\
\hline
Average & 91.4 & 86.3 \\
\end{tabular}
\end{table}

We performed the aforementioned cross-validation technique and obtained the best classifier, which is an RBF kernel with a $\gamma$ value of 0.00001 and a $C$ value of 1. We tested it on the test set, and the results are shown in Table \ref{rtsAcTT}. The accuracies when jointly trained perform relatively well as a whole, however some operating points can result in difficult to classify examples. We kept the operating points separate to observe how well the universal classifier does on each particular case and then obtained the average accuracy over the whole test set to demonstrate overall performance. 

\subsubsection{Inference with trustworthy PMUs}

We investigated the performance of the proposed method when only a small subset of PMUs are used for classification. We created a small subset of PMUs to choose from located at buses: 118, 121, 218, 221, 223, 318, 321, 323, 325. It turned out that using a smaller subset of PMUs does not substantially degrade performance if the subset is properly chosen. Among the assumed PMU locations, we selected a PMU to be allowed in the trusted subset only if they were immediately adjacent to a PCC in the network. As a result we can choose a handful of desired PMUs to be used. Out of these PMUs, for this experiment we only selected either two or three to be secure, then we trained and tested on the smaller subset. Table \ref{rtsAccuracies} illustrates the results of this experiment, whereby Class 1 represents a stable reconnection and Class 0 represents an unstable reconnection.

\begin{table}[h]
\renewcommand{\arraystretch}{1.3}
\caption{Accuracies for RTS-96 operating points with subsets of trusted PMUs}
\label{rtsAccuracies}
\centering
\begin{tabular}{c c c}
PMU location [bus number] & Class 1 accuracy [\%] & Class 0 accuracy [\%]\\
\hline
118, 318 & 92.7 & 87.6 \\
\hline
118, 321 & 92.8 & 87.6 \\
\hline
121, 318 & 92.4 & 87.5 \\
\hline
118, 121 & 90.1 & 85.6 \\
\hline
323, 325 & 92.2 & 86.8 \\
\hline
218, 321, 325 & 93.4 & 87.2 \\
\hline
221, 223, 323 & 94.1 & 86.6 \\
\hline
121, 218, 318 & 93.2 & 87.3 \\
\hline
118, 121, 218 & 90.2 & 86.0 \\
\hline
318, 323, 325 & 94.3 & 86.4 \\
\end{tabular}
\end{table}

The main reason for obtaining better results with limited PMUs in some test cases is due to the exclusion of PMUs that are either adding noise to the classifier or not providing relevant information. A higher number of features leads to the need for more training data to create an adequate classifier. If we use PMUs that do not provide useful information, building the classifier becomes difficult with limited training data. We observe that it may not be feasible to produce large quantities of training data which can lead to better results from subsets of PMUs rather than the entire set.  This is shown in randomly chosen subsets in Table \ref{rtsAccuracies} for RTS-96 as well as in the following section for the Poland case.

The above results suggest that the proposed method can be adjusted to be resilient to potential cyber attacks that may manipulate part of PMU data. In the event that the integrity of PMU measurement data is not fully guaranteed due to cyber threats\cite{pmuCyberSecurity}, we cannot rely on the classifier processing the full set of PMU measurements. To effectively handle such a case, we can prioritize protection of a certain small subset of PMUs such that the integrity of their measurements can be strongly guaranteed even in the presence of cyber adversaries. Our results imply that if the trusted subset is properly chosen, the classifier can perform with high accuracy based on the trusted PMU measurements.

\subsection{Poland Network}
For the Poland case we created twenty-four different operating points and generated roughly 240 different initial conditions total. On top of the steady state diversity implemented, we obtained data from 50 reconnection points spanning randomly between 40-55 seconds from each initial condition to implement more temporal diversity. We incorporated a protective scheme by adding overcurrent relays on each transmission line, as well as undervoltage and underfrequency relays on each bus. We allowed relay operation to trip lines, shed load, or disconnect generators. The overcurrent relays were set based upon the transmission line limits from the original test case. Table \ref{tableOCR} provides an overview on the relay configuration. 

\begin{table}[ht]
	\renewcommand{\arraystretch}{1.3}
	\caption{Overcurrent relay configuration}
	\label{tableOCR}
	\centering
	\begin{tabular}{c c c c c c}
		Point & Pickup [\%] & Trip time [sec.] \\
		\hline
		1 & 100 & 5 \\
		\hline
		2 & 125 & 0.2 \\
		\hline
		3 & 137.5 & 0.15 \\
		\hline
		4 & 150 & 0.1 \\
		\hline
	\end{tabular}
\end{table}

Underfrequency and undervoltage relays were used for bus and generator monitoring and protection. Setting the voltage thresholds is straightforward given the baseline variability of voltages for each bus. Frequency variability is more challenging to set up without obtaining more information from the operation of a large network. Thus, we grouped buses with similar frequency response and introduced different frequency threshold points throughout the system. As a result, load shedding and generator tripping due to underfrequency events allowed for heterogeneous disconnection, generally a more accurate depiction of system survival in a real case.  Synthesized time-dial points for underfrequency bus relays were setup as shown in Table \ref{load/genRelays}, depicted by rows (LS). For generator relays, a random value ($y = \{1, 2, 3, 4\}$) was chosen and scaled for the time-dial points shown in Table \ref{load/genRelays}, depicted by rows (GR).

\begin{table}[ht]
	\renewcommand{\arraystretch}{1.3}
	\caption{Undervoltage/underfrequency load shedding (LS) and under/over frequency generator (GR) relay configurations}
	\label{load/genRelays}
	\centering
	\begin{tabular}{c c c c c}
		Point & Pickup volt.~[p.u.] & Trip [sec.] & Pickup freq.~[Hz.] & Trip [sec.]\\
		\hline
		LS 1 & 0.92 & 5 & 49.5 & 5, 4, 3, 2 \\
		\hline
		LS 2 & 0.88 & 0.5 & 49 & 2, 1.5, 1, 0.5 \\
		\hline
		LS 3 & 0.75 & 0.2 & 48.5 & 1, 0.75, 0.5, 0.25 \\
		\hline
		GR 1 & - & - & 48.5, 51.5 & $y$ \\
		\hline
		GR 2 & - & - & 47.5, 52.5 & $y/2$ \\		
		\hline
		GR 3 & - & - & 46, 54 & $y/4$ \\		
		\hline		
	\end{tabular}
\end{table}

\begin{table}[ht]
\renewcommand{\arraystretch}{1.2}
\caption{Baseline Poland network accuracies}
\label{baselinePolAcs}
\centering
\begin{tabular}{c c c}
PMU location [bus number] & Class 1 accuracy [\%] & Class 0 accuracy [\%]\\
\hline
Unseen Operating Point Case & 94.4\%  &96.0\% \\
\end{tabular}
\end{table}

\begin{table*}[t]
	\renewcommand{\arraystretch}{1.3}
	\caption{Unseen operating point case accuracies for Poland network with subsets of trusted PMUs}
	\label{sepPolAc}
	\centering
	\begin{tabular}{c c c c c c c}
		\hline
		PMU location [bus number] & Class 1 accuracy [\%] & Class 0 accuracy [\%]\\
		\hline
        2218, 171, 118, 335, 2249, 214, 126, 139, 125, 303, 174, 2226, 186, 1607, 165, 1761 & 94.3\% & 95.6\% \\
        \hline
        186, 2331, 315, 139, 167, 10, 2234, 2124, 225, 2218, 2226, 178, 125, 2249, 126, 1607 & 95.8\% & 95.3\% \\
        \hline
        303, 2234, 2124, 315, 225, 335, 10, 118, 140, 2226, 2218, 214 & 88.4\% & 96.3\% \\
        \hline
        2218, 140, 174, 126, 125, 118, 2234, 171, 2124, 15, 167, 139 & 94.6\% & 95.6\% \\
        \hline
        167, 139, 214, 335, 178, 2226, 315, 118 & 89.8\% & 96.1\% \\
        \hline
        174, 2249, 2218, 118, 2331, 1607, 141, 166 & 95.6\% & 95.5\% \\
        \hline
        139, 165, 2218, 2226 & 96.0\% & 95.1\% \\ 
        \hline
        127, 2249, 118, 166 & 96.3\% & 94.3\% \\
		\hline
	\end{tabular}
\end{table*}

\begin{figure}[b]
  \centering
  \includegraphics[width=0.45\textwidth]{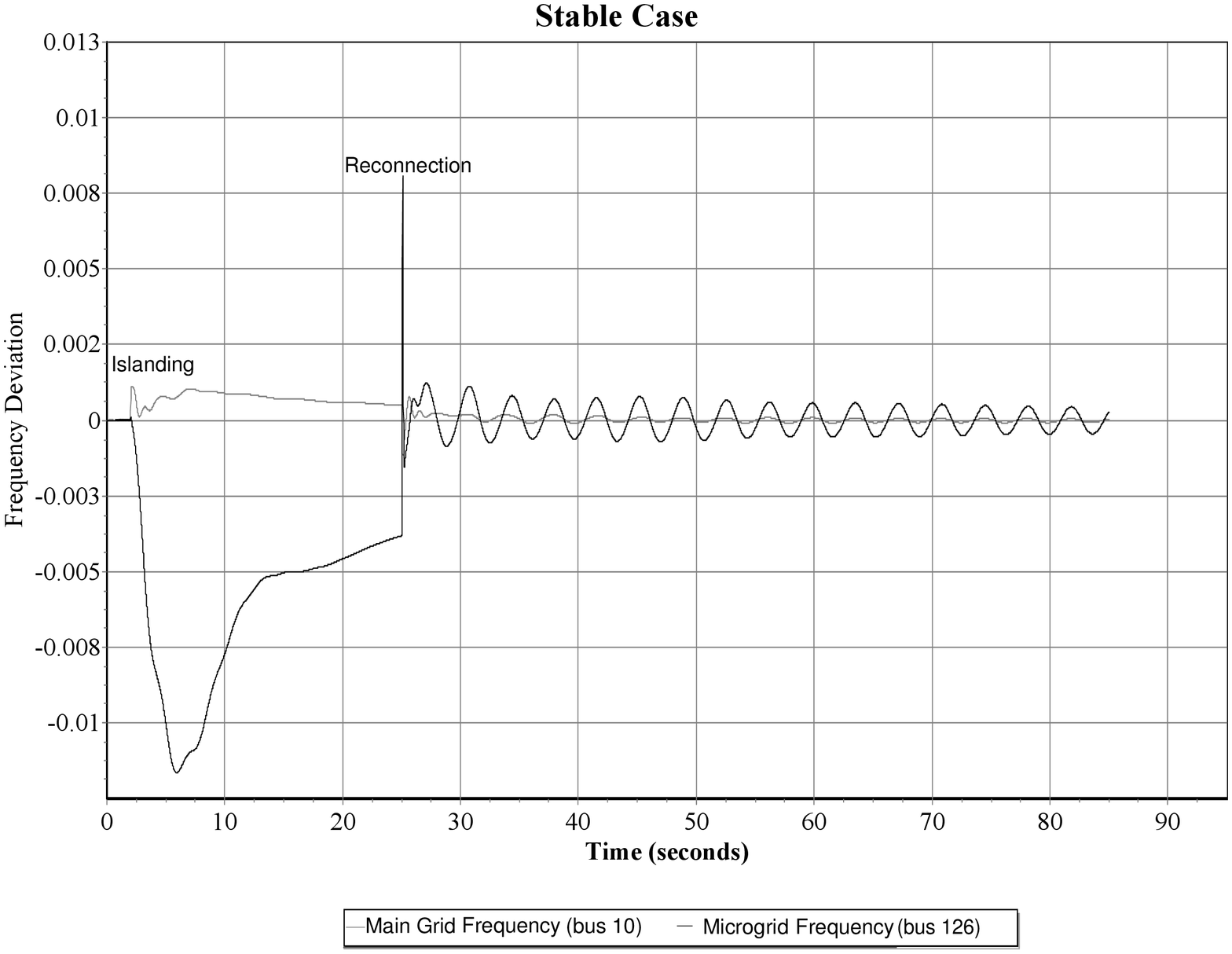}
  \caption{Stable reconnection of Poland microgrid and main grid.}
  \label{polandStable}
\end{figure}

\begin{figure}[t]
  \centering
  \includegraphics[width=0.45\textwidth]{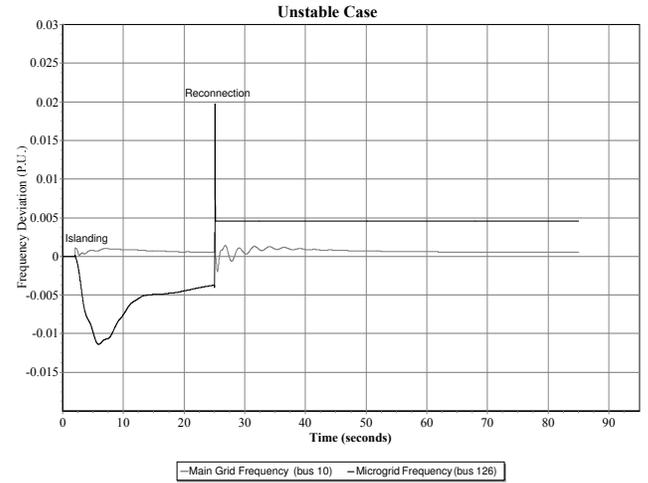}
  \caption{Unstable reconnection of Poland microgrid and main grid.}
  \label{polandUnstable}
\end{figure}

Since the Poland test case is divided by default into five zones, we solved the steady state of the case when islanding certain zones. Zone 5 was a good candidate for intentional islanding due to a low mismatch for generation and demand, as well a voltages within acceptable operating limits, thus it was selected to be the sub-network of interest in this experiment. During the dynamic simulations we islanded the sub-network at 2 seconds. We implemented a more temporal approach with respect to reconnection to capture real-time changes in the network. As a result, reconnection times ranged from 40-55 seconds for each dynamic simulation. Unlike the RTS-96 experiment, we did not assume full PMU coverage of a large scale network to begin with. We only allowed a PMU on a bus if it is immediately attached to the interconnection between the sub-network and the main grid. We were left with 30 available PMUs in the Poland network to build a feature vector. Since each PMU contains a voltage and angle measurement the dimension of the feature vector is 60 (if using the entire set of PMUs). As we stated earlier in the procedure description, the next step was to create labels based on the convergence of the case. Figures \ref{polandStable} and \ref{polandUnstable} illustrate labeling examples for stable and unstable cases, respectively.

Figs. \ref{polandStable}-\ref{polandUnstable} depict frequencies of two buses on either side of an interconnection point.  One can observe that case labeled as stable case exhibits a reconnect where the frequency signals converge to a common operating state. The unstable case shows the frequency of Bus 126 spike and immediately flat-line representing a bus trip. As described in the methodology section, if the network did not converge, it would have immediately been labeled unstable. The rules of stability in the Poland case additionally enforced that at least 2370 of the 2383 buses in the case were in service after reconnection of the island.

We partitioned the 722 different initial conditions in accordance to the two test cases described in Section III-E: multiple operating point case and unseen operating point case. For each test case, we used 10-fold cross validation together with random oversampling to learn optimal hyperparmaters and train the classifier (see\cite{Mohri2012, VanHulse2007} for details of these methods). The set aside test set was then used to determine the classifier's accuracy.

The baseline accuracies of the Poland network are seen in Table \ref{baselinePolAcs}.
The unseen operating point case represents the case in which the testing set contains data from operating points that do not exist in the training set. With the unseen operating point test, the proposed algorithm demonstrated over 90\% accuracy.  In particular, the results from the unseen operating point case suggest that our classifier can demonstrate this accuracy even when the classifier is trained based on a few operating points and tested for an \emph{unseen} operating point case.  This implies that the proposed method is scalable and suitable for use in a large-scale grid; the classifier does not have to be trained for all possible operating points, and training with a few suffices. As stated earlier for the smaller test case experiments, we also investigate the accuracy of the classifier for a scenario when the system is compromised. As a response, our classifier makes use of a trusted set of PMUs and makes predictions based on their measurements. A variety of subsets from the available PMU full set make up our possible trusted scenarios, as shown in Table \ref{sepPolAc}. The results indicate that some subsets still perform well even in the face of unknown operating points.

In the larger Poland case it seems more prevalent that decreasing the amount of features can lead to similar performance to the full set. The adoption of this control technique would bring into question whether a utility could provide enough training data, specifically the number of training examples, for the classifier. If limited training data is provided, the usage of an optimal subset of PMUs instead of the entire available set could yield adequate accuracies. Indeed it is always interesting to observe that less amount of information give similar results. It is explained in this case by considering a high dimension of the feature space. For high dimensional feature vectors, it is difficult to learn an accurate classifier with limited amount of training data. Utilities with the ability to archive and make available relatively large amounts of training data could still make use of a a large set of PMUs, if available, and potentially observe higher accuracies with respect to the quantity of training examples provided in these experiments.  

\section{Conclusion}

This paper presents a machine learning approach for the prediction of stable reconnections of a power system sub-network. The proposed approach leverages a power system dynamics simulator to generate synthetic, yet realistic in terms of size, training examples that are subsequently employed to train a classifier. The interactions between power system dynamics and protection mechanisms are complex, and the exact derivation of an optimal control strategy is not always feasible. However, as demonstrated in this paper, a machine learning approach can be useful to capture many unintuitive behaviors and make predictions in real-time based on PMU measurements. Future improvements on the training aspect may be necessary as the procedure to build said classifier is relatively sophisticated and requires in depth knowledge. The method may not be directly usable by operators as a result of the necessary understanding to build a well trained classifier. The classifier was tested on a variety of cases and operating points to ensure diversity. Accuracies of approximately 90\% were observed throughout most conditions when making dynamic predictions of a given network. Existing work in literature is limited to the dynamic realization of reconnection stability, however future work may leverage said technique in a more time sensitive way. In addition, cyber attacks on PMUs in a subset may distort the classifier thus creating the need to implement techniques on verifying the authenticity of the data streams.


%


\section*{Acknowledgment}
This material is based upon work supported by the Department of Energy under Award Number DE-OE0000780.

\ifCLASSOPTIONcaptionsoff
  \newpage
\fi



%
\bibliographystyle{IEEEtran}
\bibliography{NewVersion}


%





\end{document}